\documentclass{article} 
\usepackage[final]{colm2026_conference}

\usepackage{microtype}
\usepackage{hyperref}
\usepackage{url}
\usepackage{booktabs}

\definecolor{darkblue}{rgb}{0, 0, 0.5}
\hypersetup{colorlinks=true, citecolor=darkblue, linkcolor=darkblue, urlcolor=darkblue}

\usepackage{latexsym}

\usepackage[T1]{fontenc}

\usepackage[utf8]{inputenc}

\usepackage{graphicx}

\usepackage{amsmath}
\usepackage{multirow}
\usepackage{enumitem}
\usepackage{wrapfig}

\title{SciTaRC: A Plan-Annotated Scientific Tabular QA Benchmark for Language Reasoning and Complex Computation}

\author{
\textbf{Hexuan Wang} \quad \textbf{Yaxuan Ren} \quad \textbf{Srikar Bommireddypalli} \quad \textbf{Shuxian Chen} \\[1.5mm]
\textbf{Adarsh Prabhudesai} \quad \textbf{Rongkun Zhou} \quad \textbf{Elina Baral} \quad \textbf{Philipp Koehn} \\[2.5mm]
Center for Language and Speech Processing, Johns Hopkins University \\[2mm]
\texttt{\{hwang302, phi\}@jhu.edu}
}

\begin{document}
\maketitle

\begin{abstract}
We introduce SciTaRC, an expert-authored benchmark for question answering over scientific tables that targets composite, multi-step reasoning. To enable fine-grained diagnostic analysis beyond end-task accuracy, SciTaRC pairs each question with a manually constructed reasoning plan and explicit complexity metrics. State-of-the-art models fail on at least 23\% of these questions, while highly capable open-weight models like Llama-3.3-70B collapse on 65.5\% of the benchmark. Error analysis shows that, in zero-shot settings, failures are driven primarily by question comprehension, where models misinterpret the scientific query and derive the wrong reasoning objective. To determine whether overcoming this gap is sufficient, we use the structured plans to decouple strategy formulation from execution. Surprisingly, providing oracle step-by-step plans yields only limited gains and fails to eliminate the performance gap. This reveals a substantial execution bottleneck: both natural language and code-based methods struggle to reliably carry out long-horizon computational chains over structured data. Ultimately, SciTaRC serves as a rigorous diagnostic testbed for studying both planning and execution in scientific table reasoning.\footnote{Dataset and code: \url{https://huggingface.co/datasets/JHU-CLSP/SciTaRC} and \url{https://github.com/JHU-CLSP/SciTaRC}}
\end{abstract}

\section{Introduction}
Artificial intelligence is increasingly tasked with accelerating scientific discovery, powering applications from automated literature reviews to autonomous ``AI scientists'' \citep{wang2023scientific, lu2024aiscientist}. Scientific tables are central repositories of empirical knowledge, combining heterogeneous structure, domain-specific notation, and implicit assumptions. Answering questions over scientific tables therefore requires models to simultaneously understand language, retrieve precise evidence, and execute multi-step procedures \citep{lu_scitab_2023, sui2024tablemeetsllmlarge}. While large language models (LLMs) have advanced significantly in knowledge recall and multi-step reasoning \citep{naveed_comprehensive_2024, ferrag_reasoning_2025}, traditional algorithms often remain superior for tasks requiring long-horizon execution, precise intermediate computation, or strict control flow \citep{plaat_multi-step_2025, wu2025tabulardataunderstandingllms}. This observation has motivated hybrid approaches that combine neural and symbolic reasoning for agentic AI systems \citep{wan_towards_2024, colelough_neuro-symbolic_2025}. Scientific tables provide a natural stress test for these paradigms by demanding a seamless integration of semantic understanding and logical execution.

However, existing table QA benchmarks capture only part of this challenge. Prior work has advanced evaluation through realistic formatting, multi-table reasoning, and table-text integration \citep{qiu2024tqabenchevaluatingllmsmultitable, zhu-etal-2025-tableeval, zhang-etal-2025-scitat}, but it provides limited support for \textit{composite reasoning}. In reality, scientific analysis rarely consists of isolated tasks; rather, it requires chaining retrieval, comparison, aggregation, calculation, and conditional logic within a single, coherent solution process, iterating over variable-length sets and branching on values computed earlier.

\begin{figure}[t]
\centering
\includegraphics[width=\linewidth]{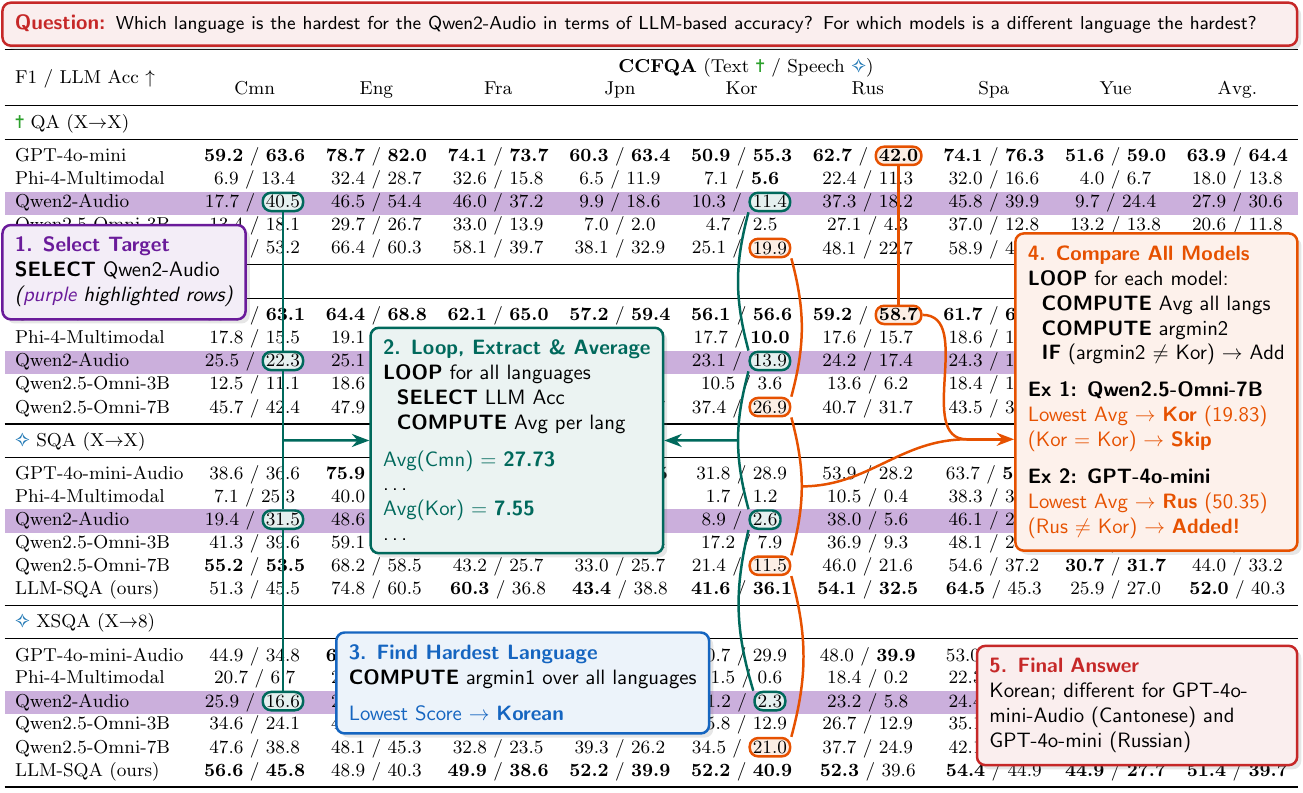}
\caption{\textbf{A representative SciTaRC task.} Instead of simple fact retrieval, the models must execute a multi-step procedure that combines targeted lookup, computation, aggregation, and conditional comparison to derive the final answer.}
\label{fig:example}
\end{figure}

To rigorously evaluate this setting, we introduce \textbf{SciTaRC} (\textbf{Sci}entific \textbf{Ta}ble QA requiring \textbf{R}easoning and \textbf{C}omputation), a benchmark of 371 expert-authored questions for question answering over scientific tables. Beyond end-task evaluation, SciTaRC offers a methodological framework for fine-grained diagnostic analysis: each question is paired with a manually written structured reasoning plan and explicit complexity annotations. Where existing benchmarks report only whether a model failed, these annotations support fine-grained analysis of the conditions under which it fails. Figure~\ref{fig:example} illustrates a representative example. Despite recent advances, state-of-the-art models such as GPT-5 achieve only 76.8\% accuracy, while highly capable open-weight models like Llama-3.3-70B-Instruct fail on over 65\% of questions.

In this paper, we detail the construction of SciTaRC, evaluate a range of state-of-the-art models, and utilize our complexity metrics and structured plans to analyze their failure modes. We find that zero-shot performance is initially constrained by a \textbf{comprehension bottleneck}, where models misinterpret the scientific query or misunderstand relevant table entities. However, providing oracle plans to remove this source of error reveals that resolving comprehension is insufficient. Instead, it exposes a severe \textbf{execution bottleneck}: models still struggle to faithfully execute long chains of computation and logic, even with the correct reasoning path. Furthermore, imposing rigid external plans on generalist models often incurs a \textbf{compliance cost} that disrupts their intrinsic reasoning. Thus, our results suggest that while planning remains an important research frontier, reliable execution over structured data is an equally critical requirement for future autonomous scientific agents.

\section{Related Work}

\textbf{Table QA Benchmarks.}
Early benchmarks such as WikiTableQuestions \citep{pasupat-liang-2015-compositional} and TabFact \citep{chen2020tabfactlargescaledatasettablebased} focused on relatively clean tables with simple reasoning patterns. More recent work has pushed toward greater complexity by targeting messy real-world formats \citep{zhu-etal-2025-tableeval}, multi-table relational reasoning \citep{qiu2024tqabenchevaluatingllmsmultitable}, and diverse analytical tasks \citep{wu2025tablebench}. Domain-specific datasets have also emerged for finance \citep{katsis-etal-2022-ait} and hierarchical structures \citep{cheng-etal-2022-hitab}. Open-domain resources such as DataBench \citep{oses-grijalba-etal-2024-question} target tabular QA over consumer, business, and survey data with typed atomic answers. Most similar to our dataset is SCITAT \citep{zhang-etal-2025-scitat}, which also addresses scientific tables, but whose questions are LLM-generated and isolate a single reasoning type (e.g., arithmetic \textit{or} lookup). Across these benchmarks, tables are cleaned into regular grids, a question is answered by a lookup or a single aggregation, and short atomic answers are scored by exact or fuzzy match. SciTaRC instead poses expert-authored questions whose answers require composite reasoning and computation, over tables kept in the raw \LaTeX{} of the papers themselves, the form an autonomous system must handle when it reads the literature directly. Because such answers are free-form and multi-part, we score them with an LLM judge validated against human annotation, and we annotate each question with a structured reasoning plan and complexity metrics that make failures diagnosable rather than merely countable.

\textbf{Reasoning Methods and Diagnostics.}
Recent surveys \citep{fang2024largelanguagemodelsllmstabular, wu2025tabulardataunderstandingllms} highlight persistent challenges in table understanding. Diagnostic studies reveal specific weaknesses: \citet{sui2024tablemeetsllmlarge} shows sensitivity to serialization formats, and \citet{akhtar-etal-2023-exploring} demonstrate inconsistency in numerical reasoning. Methodological advances include multi-agent frameworks \citep{bai-etal-2025-maple, Ma_2025}, specialized reasoning architectures \citep{xiong2025tablereasoneradvancingtablereasoning}, and inference-time scaling \citep{yang-etal-2025-table}. SciTaRC complements this work by providing complexity metrics, a diagnostic testbed, and a systematic analysis of failure modes on scientific tables.

\textbf{Neuro-Symbolic Approaches.}
Program-based reasoning methods like Program-of-Thought \citep{chen2023programthoughtspromptingdisentangling} and PAL \citep{gao2023palprogramaidedlanguagemodels} generate executable code to disentangle computation from reasoning. DACO \citep{wudaco_2024} advances code-based data analysis, while recent work explores LLM-symbolic integration \citep{kulkarni-etal-2025-llm}. Despite these advances, questions remain about whether program-based methods can handle the heterogeneous structures of scientific tables in zero-shot settings. SciTaRC addresses this through manually written structured plans that enable systematic decomposition of planning failures versus execution failures---a distinction not possible with existing benchmarks.

\section{Dataset}
Our dataset is based on scientific papers sourced from recent arXiv preprints, mainly in the domain of artificial intelligence. We consolidated the downloaded \LaTeX\ sources for each paper into a single file, deliberately retaining the raw \LaTeX\ format to reflect the authentic, heterogeneous structure of real-world scientific papers, so that the input matches what a system reading the literature directly would receive.

\subsection{Question Creation}
For each paper, a question author selected 1 to 3 tables, wrote a question, and provided the answer. Each dataset instance includes the arXiv paper ID, question text, gold answer, the \LaTeX\ sources for the relevant tables, all tables in the paper, and the full text. The question creators (all authors of this paper) range from undergraduate students to a professor in natural language processing. Questions were designed to be unambiguously answerable using only the tables, captions, and general domain knowledge. Each question was subsequently verified by two reviewers other than its creator. Some questions were too complex for manual calculation. In those cases, short ad-hoc scripts were written to compute the correct gold answer. These scripts were written in a mix of programming languages and are not included in the released dataset.

\subsection{Evaluation Protocol}
Because answers are free-form and often consist of multiple text and numeric components, matching model generations with the gold answer is non-trivial. For instance, the gold answer in Figure~\ref{fig:example} is: \textit{``Korean; different for GPT-4o-mini-Audio (Cantonese) and GPT-4o-mini (Russian)''}. However, a perfectly correct model might respond with full, descriptive sentences rather than a concise list. To comprehensively assess performance despite formatting variations, we employ two metrics:

\textbf{Exact Match (EM):} We include Exact Match as a traditional TableQA baseline, requiring strict string equivalence. While standard for simple factoid lookups, EM is fundamentally inadequate for the composite, free-form answers required by SciTaRC. It heavily penalizes models for minor formatting differences, stylistic choices, or extra words despite correct underlying reasoning, thereby measuring superficial formatting brittleness rather than true semantic or mathematical reasoning failures.

\textbf{LLM-as-a-Judge:} To effectively capture semantic and logical correctness, we adopt an LLM-as-a-Judge protocol \citep{zheng2023judgingllmasajudgemtbenchchatbot} as our primary metric. We prompt Llama-3.3-70B-Instruct \citep{grattafiori2024llama3herdmodels} with the question, the model's prediction, and the gold answer (exact prompt in Appendix~\ref{sec:eval_prompt}). To ensure this automated judge is reliable, we manually annotated a subset of 245 questions. The LLM judge demonstrated high alignment with human evaluations, achieving a 97.6\% agreement rate for GPT-4o outputs and 95.5\% for Llama-3-8B-Instruct outputs.


\subsection{Structured Reasoning Plans}
\label{sec:pseudo_code}

To systematically diagnose whether models fail at strategy formulation (planning) or stepwise implementation (execution), we pair each question with a manually created reasoning plan (see the visual overview in Figure~\ref{fig:example}, and Appendix~\ref{sec:appendix_example_plan} for the complete plan and further examples). Expressing these plans in \textbf{structured natural language} rather than a complex programming language removes strict syntax barriers. This deliberate design reduces the risk that observed failures reflect compilation or parsing errors rather than flawed logical execution.

To ensure consistency and eliminate stylistic confounders, the syntax is standardized around five core operations: \textbf{SELECT} (retrieves table values), \textbf{COMPUTE} (executes mathematical or comparative operations), \textbf{IF .. ELSE} (evaluates conditional logic), \textbf{LOOP} (iterates over lists), and \textbf{RETURN} (outputs the final result). This framework mimics natural problem-solving. Operations can be applied directly over lists without explicit loops, with nested logic indicated simply by indentation. To maintain conversational flow, variables passed sequentially between statements remain implicit. Explicit naming (e.g., \texttt{overall\_mean}) is strictly reserved for tracking concurrent intermediate results across long execution horizons.

\subsection{Quantifying Complexity}
\label{sec:metrics}
To quantify the challenges of SciTaRC, we define metrics characterizing difficulty along two dimensions. Table~\ref{tab:stats} summarizes the dataset statistics across these dimensions.
\begin{table}[ht]
\centering
\small
\renewcommand{\arraystretch}{1.1}
\begin{tabular*}{\linewidth}{@{\extracolsep{\fill}} l ccc cccc @{}}
\toprule
& \multicolumn{3}{c}{\textbf{Input Complexity}} & \multicolumn{4}{c}{\textbf{Reasoning Complexity}} \\
\cmidrule(lr){2-4} \cmidrule(lr){5-8}
\textbf{Metric} & $S_{tab}$ & $S_{cell}$ & $S_{tok}$ & $L_{plan}$ & $I_{retr}$ & $I_{calc}$ & $C_{flow}$ \\
\midrule
Mean & 1.3 & 287.0 & 2,527 & 9.8 & 3.4 & 4.0 & 3.0 \\
Max  & 3 & 3,338 & 22,671 & 27 & 18 & 18 & 12 \\
\bottomrule
\end{tabular*}
\caption{\textbf{SciTaRC Dataset Statistics.} These metrics illustrate the composite reasoning focus over large scientific tables. High average values for plan horizon ($L_{plan}$) and table size ($S_{cell}$) highlight the requirement for high-fidelity execution across long-horizon scientific contexts.}
\label{tab:stats}
\end{table}

\textbf{Input Complexity.}
We measure the amount of information to be processed:
\begin{itemize}[leftmargin=*,noitemsep,topsep=0pt]
    \item \textbf{Context Load ($S_{tok}$):} Total token count of serialized tables, measuring context length.
    \item \textbf{Structural Size ($S_{cell}$):} Total grid cells parsed from \LaTeX{}, a proxy for the number of data items.
    \item \textbf{Input Scope ($S_{tab}$):} Number of tables required, measuring cross-table linking needs.
\end{itemize}

\textbf{Reasoning Complexity.}
We quantify algorithmic hardness via ground-truth plans:
\begin{itemize}[leftmargin=*,noitemsep,topsep=0pt]
    \item \textbf{Calculation Intensity ($I_{calc}$):} Count of \texttt{COMPUTE} operations, measuring arithmetic burden.
    \item \textbf{Retrieval Demand ($I_{retr}$):} Count of \texttt{SELECT} operations, measuring retrieval frequency.
    \item \textbf{Plan Horizon ($L_{plan}$):} Total instruction count, serving as a proxy for reasoning depth.
    \item \textbf{Control Flow ($C_{flow}$):} Composite score $C_{flow} = N_{loop} + N_{if} + D_{max}$, where $N_{loop}$ and $N_{if}$ count iterative and conditional blocks, and $D_{max}$ represents maximum nesting depth.
\end{itemize}

\section{Benchmarking State-of-the-Art}
\label{sec:benchmark}

\subsection{Benchmark Setup}
We evaluate 24 models spanning two primary categories: \textbf{Proprietary Models} (e.g., GPT-5, Grok-4.1) and \textbf{Open-Weight Models}. The latter includes both leading generalist models (e.g., DeepSeek-V3.2~\citep{deepseekai2025deepseekv32pushingfrontieropen}, Llama 3.3~\citep{grattafiori2024llama3herdmodels}, Qwen 3~\citep{yang2025qwen3technicalreport}) and code-optimized specialists (e.g., Qwen-Coder~\citep{hui2024qwen25codertechnicalreport}).

We employ two zero-shot prompting paradigms tailored to the strengths of each model: \textbf{Chain-of-Thought (CoT)}~\citep{wei2023chainofthoughtpromptingelicitsreasoning, kojima2023largelanguagemodelszeroshot} for generalist models to encourage step-by-step natural language reasoning, and \textbf{Program-of-Thought (PoT)}~\citep{chen2023programthoughtspromptingdisentangling} for code-optimized models and select generalists (e.g., DeepSeek-V3.2, Claude 4.5 Sonnet) to generate executable Python code. All prompts are provided in Appendix~\ref{sec:inference_prompts}.

\begin{figure}[t]
    \centering
    \includegraphics[width=\linewidth]{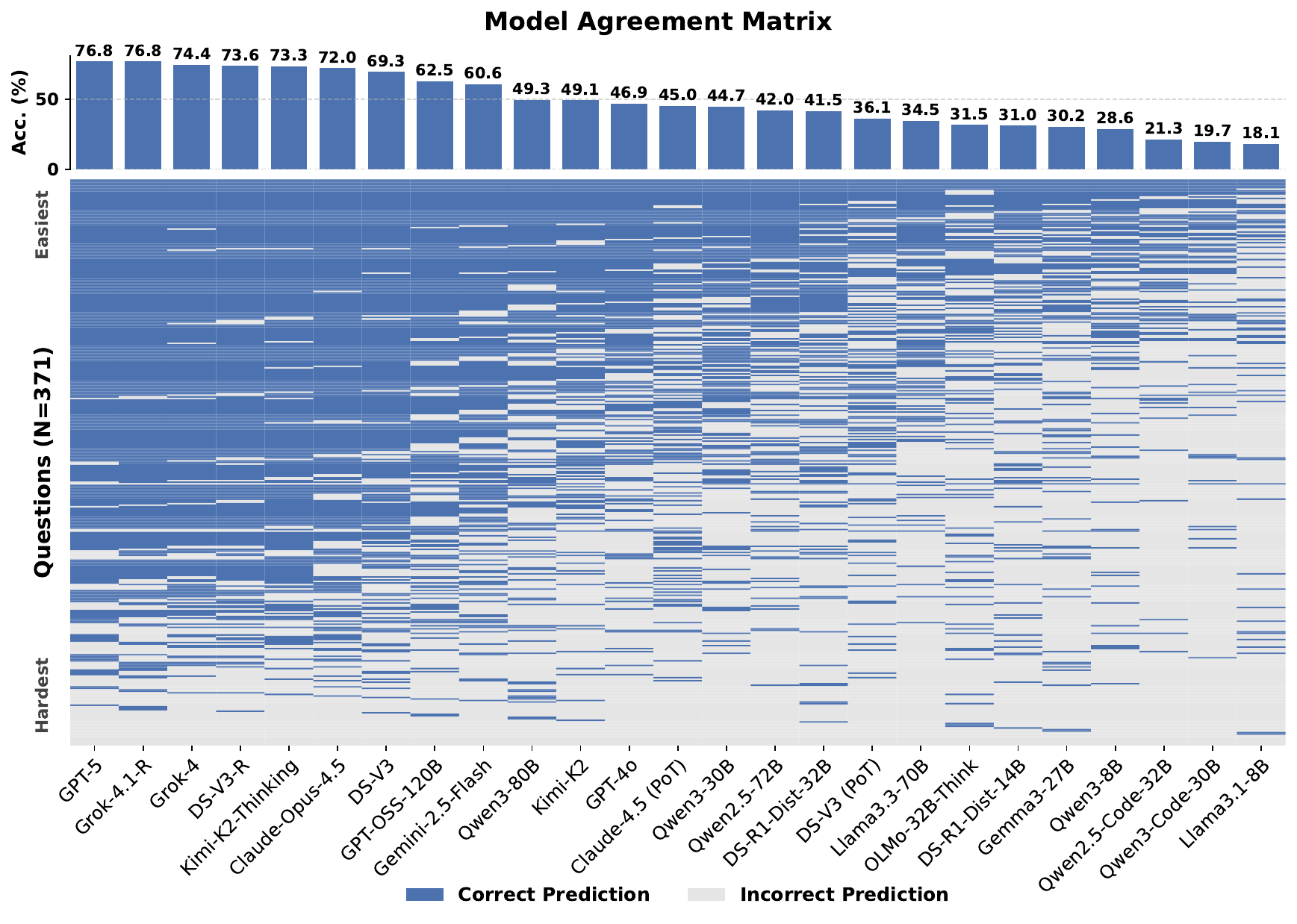}
    \caption{\textbf{Model Agreement Matrix.} Rows represent questions sorted by difficulty, defined as the total number of models that answered correctly (easiest at top). Columns represent models sorted by accuracy. \textbf{Blue} indicates a correct prediction. }
    \label{fig:heatmap}
\end{figure}

\subsection{Zero-Shot Benchmark Performance}
\label{sec:main_results}

\begin{table}[t]
\centering
\small
\renewcommand{\arraystretch}{1.1} 
\begin{tabular*}{\linewidth}{@{\extracolsep{\fill}} l l c c c c @{}}
\toprule
\textbf{Model Family} & \textbf{Model Name} & \textbf{Method} & \textbf{Size} & \textbf{LLM-Judge (\%)} & \textbf{EM (\%)} \\
\midrule
\multicolumn{6}{c}{\textit{\textbf{Proprietary Models}}} \\
\midrule
OpenAI & GPT-5 & CoT & - & \textbf{76.8} & 22.1 \\
xAI & Grok-4.1-fast-reasoning & CoT & - & \textbf{76.8} & 2.7 \\
xAI & Grok-4 & CoT & - & 74.4 & 34.0 \\
Moonshot & Kimi-K2-Thinking & CoT & - & 73.3 & 24.8 \\
Anthropic & Claude-Opus-4.5 & CoT & - & 72.0 & 20.2 \\
Google & Gemini-2.5-Flash & CoT & - & 60.6 & 31.0 \\
Moonshot & Kimi-K2 & CoT & - & 49.1 & 26.2 \\
OpenAI & GPT-4o & CoT & - & 46.9 & 10.8 \\
Anthropic & Claude-Sonnet-4.5 & PoT & - & 45.0 & 0.5 \\
\midrule
\multicolumn{6}{c}{\textit{\textbf{Open-Weight Models}}} \\
\midrule
DeepSeek & DeepSeek-V3.2 (thinking) & CoT & 685B & \textbf{73.6} & \textbf{32.1} \\
DeepSeek & DeepSeek-V3.2 (non-thinking) & CoT & 685B & 69.3 & 11.6 \\
OpenAI & GPT-OSS-120B & CoT & 120B & 62.5 & 0.3 \\
Qwen & Qwen3-Next-80B-A3B & CoT & 81B & 49.3 & 12.7 \\
Qwen & Qwen3-30B-A3B-2507 & CoT & 30B & 44.7 & 8.1 \\
Qwen & Qwen2.5-72B-Instruct & CoT & 73B & 42.0 & 4.9 \\
DeepSeek & DS-R1-Distill-Qwen-32B & CoT & 32B & 41.5 & 10.5 \\
DeepSeek & DeepSeek-V3.2 (non-thinking) & PoT & 685B & 36.1 & 0.8 \\
Meta & Llama-3.3-70B-Instruct & CoT & 71B & 34.5 & 5.1 \\ 
AllenAI & OLMo-3.1-32B-Think & CoT & 32B & 31.5 & 1.6 \\
DeepSeek & DS-R1-Distill-Qwen-14B & CoT & 14B & 31.0 & 7.0 \\
Google & Gemma-3-27B-IT & CoT & 27B & 30.2 & 0.8 \\
Qwen & Qwen-3-8B & CoT & 8B & 28.6 & 0.0 \\
Qwen & Qwen-2.5-Coder-32B & PoT & 32B & 21.3 & 6.7 \\
Qwen & Qwen-3-Coder-30B & PoT & 30B & 19.7 & 6.7 \\
Meta & Llama-3.1-8B-Instruct & CoT & 8B & 18.1 & 0.0 \\
\bottomrule
\end{tabular*}
\caption{\textbf{SciTaRC Main Leaderboard.} Models are grouped by access level and sorted by primary accuracy. All evaluations are zero-shot, using CoT by default and PoT for code-optimized models. Our primary metric, \textbf{LLM-Judge (\%)}, evaluates semantic and logical correctness; \textbf{EM (\%)} measures strict string equivalence as a traditional TableQA baseline. Best results per category are \textbf{bolded}.}
\label{tab:main_results}
\vspace{-1mm} 
\end{table}

Table~\ref{tab:main_results} summarizes the zero-shot performance across all models, while Figure~\ref{fig:heatmap} visualizes agreement at the question level. We observe three primary trends:

\textbf{Reasoning Strategies Amplify Capabilities.}
Models trained with reinforcement learning~\citep{deepseekai2025deepseekr1incentivizingreasoningcapability} for reasoning significantly boost performance, enabling models to outperform larger standard baselines. For instance, Kimi-K2-Thinking (73.3\%) exceeds its base model Kimi-K2 (49.1\%) by over 24 percentage points. This capability allows efficient scaling: the 32B DeepSeek-R1-Distill reasoning model (41.5\%) scores higher than the 71B Llama-3.3 model (34.5\%).

\textbf{Model Scale Matters, but Capabilities Vary.}
While larger models generally achieve higher overall accuracy, their capabilities are not a strict superset of weaker models. As shown in Figure~\ref{fig:heatmap}, smaller models occasionally solve ``hard'' questions that frontier models miss. This stochasticity suggests that reasoning is not a single monotonic skill; different architectures may find different types of logic easier, even if they lack overall capacity.

\textbf{Zero-Shot Code Execution Struggles.}
Contrary to the expectation that tables are best handled by code, Program-of-Thought (PoT) performs substantially worse than natural language reasoning. Strong generalist models suffer massive degradation when forced to write code—DeepSeek-V3.2 drops from 69.3\% (CoT) to 36.1\% (PoT)—and dedicated code specialists like Qwen-2.5-Coder achieve only 21.3\%. Code appears brittle when applied to the ad-hoc, heterogeneous structures of scientific tables, whereas natural language is more robust to interpret the noise.

\subsection{Analysis of Difficulty Factors}
\label{sec:complexity_analysis}

To understand the mechanisms behind model failure, we first analyze performance against the complexity metrics defined in Section~\ref{sec:metrics}. We focus on two dimensions: the scale of the input data and the complexity of the required reasoning.

\textbf{Input Scale Exposes Fragility.}
Performance generally degrades as the structural size of the table ($S_{cell}$) increases (see Appendix~\ref{sec:appendix_scale} for detailed visualizations). This downward trend is consistent across both natural language (CoT) and code generation (PoT) paradigms. However, the data reveals a sharp divide in model robustness. Frontier models exhibit remarkable stability: Grok-4 (+0.1\%) and DS-V3 (-2.0\%) maintain their accuracy even as tables exceed 400 cells. In contrast, smaller models suffer catastrophic collapse due to context pressure, with Gemma-3-27B dropping by 41.4 percentage points. We observe similar patterns for Context Load ($S_{tok}$) and Input Scope ($S_{tab}$).

\begin{figure*}[h]
    \centering
    \includegraphics[width=\linewidth]{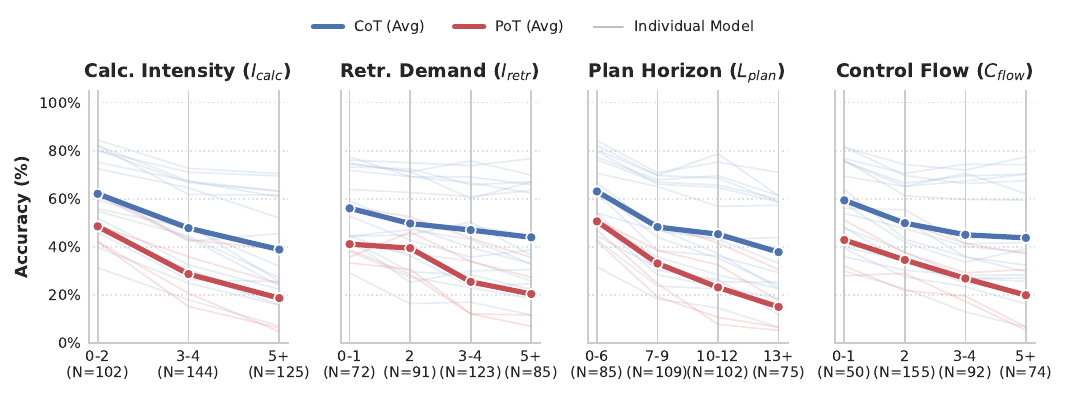}
    \caption{\textbf{Impact of Reasoning Complexity.} We measure accuracy against four metrics: Calculation Intensity ($I_{calc}$), Retrieval Demand ($I_{retr}$), Plan Horizon ($L_{plan}$), and Control Flow ($C_{flow}$). Across all dimensions, increased complexity correlates with lower performance.}
    \label{fig:reasoning_complexity}
\end{figure*}

\textbf{Impact of Reasoning Complexity.}
Beyond data scale, we examine how the algorithmic requirements of the question affect performance (Figure~\ref{fig:reasoning_complexity}). We observe a \textbf{consistent negative trend} across all metrics, but the drivers of failure differ.
\begin{itemize}[leftmargin=*,noitemsep,topsep=2pt]
    \item \textbf{Plan Horizon is a Universal Bottleneck ($L_{plan}$):} Multi-step reasoning proves uniquely difficult for \textit{all} systems. Even robust frontier models like DeepSeek-V3.2 (thinking) drop by 13.4\% when moving to long-horizon questions, and the average system degrades by 25.2\% to 35.5\% depending on the method (CoT or PoT). This suggests that maintaining execution fidelity across long sequences remains a persistent challenge across the board.
    \item \textbf{Code Generation Degrades with Complexity ($I_{calc}$):} Contrary to the expectation that code handles math better, Program-of-Thought (PoT) degrades faster than natural language on calculation-heavy tasks. 
    PoT accuracy drops by 29.8\% (vs 23.2\% for CoT), likely because generating the correct syntax for long operation chains zero-shot is more prone to error than natural language reasoning.
    \item \textbf{Complex Logic Separates Reasoning Models ($C_{flow}$):} Questions with nested loops or conditionals reveal a stark divergence. While standard models like GPT-4o suffer significant degradation (-29.5\%), reasoning-optimized models like Kimi-Thinking and DeepSeek-V3.2 remain remarkably robust (+0.9\% and -2.6\% respectively), indicating their specific training enables them to handle structural logic that confuses generalist models.
\end{itemize}

\section{Error Analysis: The Comprehension Bottleneck}
\label{sec:error_analysis}

To understand the widespread zero-shot failures observed in Section~\ref{sec:main_results}, we conducted a qualitative analysis of 98 incorrect outputs from DeepSeek-V3.2 (thinking). We categorize these failures using a four-class taxonomy:

\begin{itemize}[leftmargin=*,noitemsep,topsep=2pt]
\item \textbf{Comprehension}: Misunderstanding the question intent, constraints, or table attributes, leading to the wrong objective.
\item \textbf{Localization}: Selecting incorrect rows, columns, or values despite a correct objective.
\item \textbf{Calculation}: Incorrect numeric result given the correct evidence and operation.
\item \textbf{Memory}: Failing to retain and apply intermediate results across multi-step plans.
\end{itemize}

\begin{wraptable}{r}{0.35\textwidth}
\centering
\small
\begin{tabular}{l r}
\toprule
\textbf{Category} & \textbf{Ratio}\\
\midrule
Comprehension & 73\%\\
Calculation & 17\%\\
Localization & 7\%\\
Memory & 3\%\\
\bottomrule
\end{tabular}
\caption{Error distribution.}
\label{tab:error-categories}
\end{wraptable}

Each failure is assigned based on its earliest unrecoverable point. Table~\ref{tab:error-categories} shows that \textbf{Comprehension} errors dominate (73\%), revealing that models struggle most with semantic grounding—translating complex scientific queries into correct operational objectives. Conversely, the low \textbf{Localization} rate (7\%) demonstrates that parsing raw \LaTeX\ structure is not the primary hurdle. We note that comprehension is the first step, and the model may have made subsequent errors which we did not count.


\section{Disentangling Reasoning: The Planning Ablation}
\label{sec:planning}

Section~\ref{sec:error_analysis} revealed that zero-shot models fail predominantly at comprehension—translating queries into actionable objectives. This raises a critical hypothesis: if we completely bypass this barrier by providing an explicit, structured reasoning strategy, the performance gap should theoretically close. To test this, we systematically decouple strategy formulation from mechanical execution via a controlled planning ablation.

\subsection{Experimental Design}
\label{sec:planning_setup}
We evaluate four representative models (Qwen2.5-Coder-32B, DS-R1-Distill-32B, Qwen3-30B, Qwen3-80B) across three settings (prompts in Appendix~\ref{sec:inference_prompts}): (1)~\textbf{Direct Inference ($S_{direct}$)}, the baseline zero-shot condition solving the problem in a single pass; (2)~\textbf{Autonomous Planning ($S_{auto}$)}, a two-stage pipeline where the model generates a plan (Section~\ref{sec:pseudo_code}) and then executes it in a fresh context; and (3)~\textbf{Oracle Planning ($S_{oracle}$)}, which injects the ground-truth plan into the execution context. Crucially, identical generation prompts for $S_{auto}$ and $S_{oracle}$ isolate the impact of plan quality.

To analyze planning efficacy, we define the \textbf{Gain Curve}. We rank questions $q_1,\dots,q_N$ by baseline difficulty under $S_{\text{direct}}$. Let $y_i\in\{0,1\}$ denote correctness for a given model on $q_i$. We define the Success Probability as a centered sliding-window average
$P_{\text{success}}(i)=\frac{1}{|W_i|}\sum_{j\in W_i}y_j$,
where $W_i=\{j:\max(1,i-\lfloor w/2\rfloor)\le j\le\min(N,i+\lfloor w/2\rfloor)\}$ and $w=50$.
This filters stochastic noise to visualize robust trends across the difficulty spectrum.

\subsection{Results: The Execution Bottleneck}

\begin{table}[t]
\centering
\small 
\renewcommand{\arraystretch}{1.1} 
\begin{tabular*}{\linewidth}{@{\extracolsep{\fill}} l ccc | ccc @{}}
\toprule
\multirow{2}{*}{\textbf{Model}} & \multicolumn{3}{c}{\textbf{Accuracy (\%)}} & \multicolumn{3}{c}{\textbf{Diagnostics}} \\
\cmidrule(lr){2-4} \cmidrule(lr){5-7}
& \textbf{Direct} & \textbf{Auto} & \textbf{Oracle} & $\Delta_{\text{Auto}}$ & $\Delta_{\text{Max}}$ & $\Delta_{\text{Plan}}$ \\
\midrule
Qwen2.5-Coder-32B & 21.3 & 23.2 & \textbf{27.5} & \textcolor{teal}{+1.9} & \textcolor{teal}{+6.2} & 4.3 \\
DS-R1-Dist-32B    & 41.5 & 36.1 & \textbf{45.8} & \textcolor{red}{-5.4} & \textcolor{teal}{+4.3} & 9.7 \\
Qwen3-30B         & 44.7 & 34.8 & \textbf{45.3} & \textcolor{red}{-10.0} & \textcolor{teal}{+0.6} & 10.5 \\
Qwen3-80B         & \textbf{49.3} & 41.8 & 49.1 & \textcolor{red}{-7.5} & \textcolor{red}{-0.2} & 7.3 \\
\bottomrule
\end{tabular*}
\caption{\textbf{Impact of Planning and Upper Bounds.} \textbf{Diagnostics}: $\Delta_{\text{Auto}}$ denotes the gain from autonomous planning (Auto $-$ Direct). $\Delta_{\text{Max}}$ denotes the absolute potential gain when the comprehension bottleneck is eliminated (Oracle $-$ Direct). $\Delta_{\text{Plan}}$ represents the performance gap caused by flawed strategy formulation (Oracle $-$ Auto). Even with perfect Oracle guidance, models hit a hard ceiling at $\sim$49\% accuracy.}
\label{tab:planning_results}
\end{table}

\begin{figure}[ht]
    \centering
    \includegraphics[width=\linewidth]{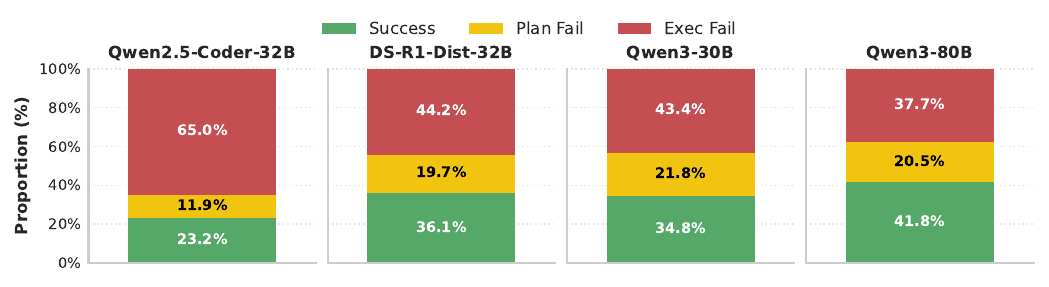}
    \caption{\textbf{Decomposition of Failure Modes.} We classify failure modes by testing if the Oracle plan rescues autonomous ($S_{auto}$) failures. \textbf{Plan Failure (Yellow)}: Failed autonomously but succeeded with an Oracle plan (flawed strategy). \textbf{Execution Failure (Red)}: Failed under both autonomous and Oracle conditions (unrecoverable by planning). The dominant red bars confirm that execution capacity, not strategy, is the primary bottleneck.}
    \label{fig:bottleneck}
\end{figure}

\begin{figure}[ht]
    \centering
    \includegraphics[width=\linewidth]{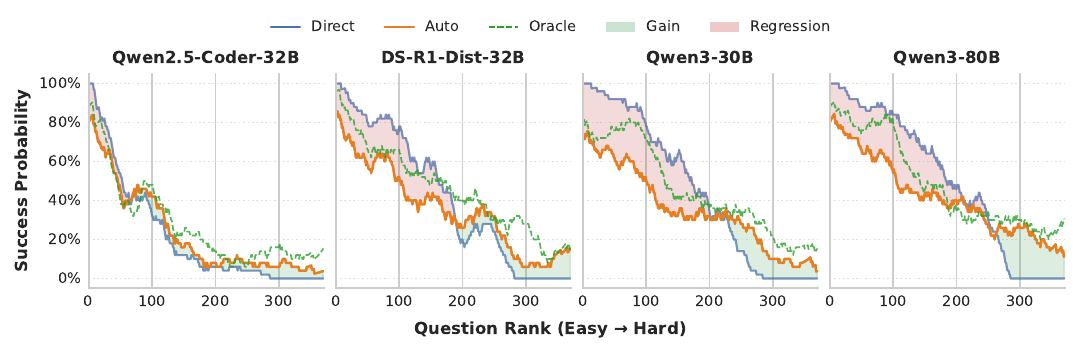}
    \caption{\textbf{Performance Gain Curves.} We plot success probability across the difficulty spectrum (see Appendix~\ref{sec:appendix_planning} for question-level changes). \textbf{Code Models (Left)} show consistent gains from planning on hard tasks. \textbf{Generalist Models (Right)} often regress on easy tasks (Red Shading) and gain on original unsolved tasks (Green Shading)}
    \label{fig:gain_curves}
\end{figure}

\textbf{Execution is the Hard Ceiling.}
Table~\ref{tab:planning_results} and Figure~\ref{fig:bottleneck} reveal that eliminating the comprehension barrier does not close the performance gap. Even when provided with a perfect Oracle plan to eliminate comprehension errors, models fail to exceed a hard ceiling of $\sim$49\% accuracy. As Figure~\ref{fig:bottleneck} visually decomposes, absolute \textbf{Execution Failure (Red)}—failing under both autonomous and Oracle conditions—dominates across all models, consistently exceeding Plan Failure (Yellow). For example, Qwen-Coder fails over 65\% of questions due to execution errors, compared to just $\sim$12\% that are recoverable from flawed planning strategies. This exposes a severe \textbf{Execution Bottleneck}: models fundamentally struggle significantly more to \textit{faithfully execute} multi-step numerical and logical operations than to \textit{formulate} the required strategy.

\textbf{The Cost of Compliance.}
Figure~\ref{fig:gain_curves} and Table~\ref{tab:planning_results} illustrate a distinct capability trade-off. Explicit planning aids the code-specialized Qwen-Coder (+1.9\% Auto, +6.2\% Oracle), confirming that structured planning reduces errors in programmatic generation. Conversely, generalist language models consistently regress (-5.4\% to -10.0\%). As visualized by the red shading in the Gain Curves, this highlights a \textbf{Compliance Cost}: forcing a generalist model to strictly follow a rigid algorithmic path actively disrupts its natural reasoning flow on simpler tasks where unguided intuition already suffices. 

Moreover, for the largest generalist model (Qwen3-80B), even perfect Oracle guidance (49.1\%) fails to surpass its zero-shot baseline (49.3\%). This suggests that an external strategy must \textit{naturally align} with a model's intrinsic reasoning mode to be effective; otherwise, the cognitive overhead of strict compliance outweighs the benefit of strategic clarity. However, for smaller architectures, the Oracle Plan remains an upper bound, revealing significant latent execution capability that autonomous planning fails to unlock.

\section{Conclusion}
We presented SciTaRC, a novel expert-authored benchmark for scientific table question answering that targets composite, multi-step reasoning over heterogeneous structured data. Beyond questions and answers, SciTaRC supplies a structured reasoning plan and complexity metrics for every instance, which is what allows failures to be diagnosed rather than merely counted. Our experiments show that this setting challenges even frontier models, which fail at least 23\% of the benchmark. Our analyses reveal a two-stage failure pattern: zero-shot models frequently misinterpret the reasoning objective, and supplying oracle plans to remove that barrier still leaves a large gap, indicating that faithful execution, not strategy formation, is the binding constraint. Isolating these limitations highlights two priorities for future architecture design. First, to resolve the compliance cost in generalist models, future systems need methods that provide strategic guidance without disrupting intrinsic reasoning. Second, the brittleness of zero-shot code generation points to the need for deeper neuro-symbolic integration, moving beyond isolated code execution. As the field advances toward autonomous AI scientists, reliable execution of complex mathematical and logical operations over structured empirical data remains a prerequisite that current systems do not meet. SciTaRC provides a rigorous testbed to measure and drive this progress.

\section*{Acknowledgments}
This work is supported in part by the Defense Advanced Research Projects Agency (DARPA) under Contract No. HR001125C0304. This work is also supported in part by an Amazon Initiative for Artificial Intelligence (AI2AI) Faculty Research Award.

\bibliography{custom_new}
\bibliographystyle{colm2026_conference}

\clearpage
\appendix

\section{LLM Usage Disclosure}
LLMs were used for generating plots, formatting LaTeX tables, and limited code drafting during development. All experimental design decisions, data construction, analysis, and final code were reviewed, verified, and finalized by the authors. LLM tools did not contribute to the scientific claims or conclusions presented in this paper. In addition, Llama-3.3-70B-Instruct serves as the automatic judge in our primary evaluation metric (Appendix~\ref{sec:eval_prompt}).

\section{Evaluation Prompt}
\label{sec:eval_prompt}

We designed the evaluation prompt to address three specific scoring challenges in scientific table reasoning:
\begin{itemize}
    \item \textbf{Stylistic Invariance:} Surface-level variations (e.g., typos, phrasing) are ignored provided the core factual content is accurate.
    \item \textbf{Numerical Tolerance:} To account for rounding differences, the protocol permits a deviation of $\pm 1$ in the final significant digit.
    \item \textbf{Strictness on Extra Information:} To avoid rewarding recall bias, the inclusion of unrequested components not present in the ground truth is penalized.
\end{itemize}

\noindent \textbf{Note on Binarization:} While the prompt utilizes a ternary scale (0.0, 0.5, 1.0) to allow the judge to explicitly categorize partial matches, we strictly binarize these scores for the final leaderboard. The ternary structure serves to isolate partially correct answers from the fully correct class; for our reported metrics, only a score of 1.0 is treated as a success (1), while 0.5 and 0.0 are treated as failures (0).

\begin{table*}[h]
\centering
\small
\renewcommand{\arraystretch}{1.1}
\begin{tabular}{p{0.95\textwidth}}
\toprule
\textbf{System Prompt} \\
\midrule
You are evaluating a table QA system. Compare the predicted answer against the ground truth answer.

\vspace{0.5em}
\textbf{SCORING RULES (follow exactly):} \\
- 1.0: CORRECT - Prediction contains the same core factual information as ground truth \\
- 0.5: PARTIALLY CORRECT - Prediction has some correct information but missing key parts \\
- 0.0: INCORRECT - Prediction is factually wrong or completely unrelated

\vspace{0.5em}
\textbf{CRITICAL REQUIREMENTS:} \\
- ALL components in ground truth MUST be mentioned in prediction \\
- If ground truth has multiple parts, prediction must include ALL parts \\
- Missing ANY component = maximum score 0.5 \\
- Adding extra components not in ground truth = maximum score 0.5 \\
- Numbers must be nearly identical - only final digit can differ by ±1

\vspace{0.5em}
\textbf{EXAMPLES:} \\
- Ground: "Japanese" | Prediction: "None. HTML2Text performs best in Japanese" -> 1.0 \\
- Ground: "Adam" | Prediction: "Adam optimizer with 0.001 learning rate" -> 1.0 \\
- Ground: "ResNet-50" | Prediction: "ResNet-50 and DenseNet-121" -> 0.5 \\
- Ground: "4.2" | Prediction: "approximately 3.8" -> 0.0

\vspace{0.5em}
\textbf{CRITICAL:} Return ONLY this exact JSON format with start/end tags: \\

[Evaluation Start] \\
{"reasoning": "Brief explanation", "score": 1.0} \\ {}
[Evaluation End]

\vspace{0.5em}
Now evaluate: \\
Question: \textbf{\{question\}} \\
Ground Truth: \textbf{\{ground\_truth\}} \\
Prediction: \textbf{\{prediction\}} \\

[Evaluation Start] \\
\bottomrule
\end{tabular}
\caption{The LLM-as-a-Judge prompt used for automatic evaluation.}
\label{tab:eval_prompt}
\end{table*}

\section{Example Questions and Reasoning Plans}
\label{sec:appendix_example_plan}

Tables~\ref{tab:appendix_examples} and~\ref{tab:appendix_examples_cont} present five complete instances spanning the complexity range of SciTaRC, each with its question, gold answer, expert-authored structured reasoning plan, and plan-derived metrics. The fourth instance is the task shown in Figure~\ref{fig:example}. Together they illustrate that question length is a poor proxy for difficulty: the second and fifth questions are stated in under thirty words, yet their plans require nested iteration over variable-length sets and branching on values computed earlier.

\begin{table*}[p]
\centering
\small
\renewcommand{\arraystretch}{1.15}
\begin{tabular}{@{}p{0.96\textwidth}@{}}
\toprule
\textbf{Q:} How much lower is the F1-score of the second-best model compared to GPT-4? \quad \textbf{A:} 0.1066 \\[1mm]
\texttt{SELECT F1-scores for all models} \\
\texttt{COMPUTE sort models by F1-score in descending order} \\
\texttt{COMPUTE difference = F1-score of GPT-4 $-$ F1-score of second-best model} \\
\texttt{RETURN difference} \\[1mm]
\textit{$L_{plan}=4$, $I_{retr}=1$, $I_{calc}=2$, $C_{flow}=0$, $S_{tab}=1$, $S_{cell}=130$} \\
\midrule
\textbf{Q:} Which zero-shot models outperform any of the fine-tuned models on any metric and language direction? \quad \textbf{A:} InternVL2.5-38B, InternVL2.5-78B, and Qwen2.5-VL-72B (on chrF++ EN-ZH) \\[1mm]
\texttt{SELECT Zero-shot Systems, all metrics, both language directions} \\
\texttt{LOOP for each zero-shot model} \\
\hspace*{4mm}\texttt{SELECT scores for current zero-shot Model} \\
\hspace*{4mm}\texttt{LOOP for each metric and language direction} \\
\hspace*{8mm}\texttt{SELECT Fine-tuned models scores} \\
\hspace*{8mm}\texttt{COMPUTE max\_ft\_score = max fine-tuned model scores} \\
\hspace*{8mm}\texttt{COMPUTE diff = zero-shot model score $-$ max\_ft\_score} \\
\hspace*{8mm}\texttt{IF diff > 0} \\
\hspace*{12mm}\texttt{COMPUTE add zero-shot model to result\_list} \\
\texttt{RETURN result\_list} \\[1mm]
\textit{$L_{plan}=10$, $I_{retr}=4$, $I_{calc}=3$, $C_{flow}=6$, $S_{tab}=1$, $S_{cell}=160$} \\
\midrule
\textbf{Q:} Considering only the models that appear in all three tables, between the Single-Domain$\rightarrow$Cross-Domain and the Cross-Domain$\rightarrow$Multi-Domain transitions, and considering all OOD ratios, how many of those model-transition-OOD-ratios yield an increase in overall F1? \quad \textbf{A:} 12 \\[1mm]
\texttt{SELECT Models present in all 3 Tables} \\
\texttt{LOOP for each Model} \\
\hspace*{4mm}\texttt{SELECT 60\% OOD Ratio from Table 1, 80\% from Table 2, and 90\% from Table 3} \\
\hspace*{4mm}\texttt{LOOP for each OOD Ratio} \\
\hspace*{8mm}\texttt{SELECT Overall F1 scores for SD, CD, and MD} \\
\hspace*{8mm}\texttt{LOOP for each Transition} \\
\hspace*{12mm}\texttt{IF Transition is "Single-Domain$\rightarrow$Cross-Domain"} \\
\hspace*{16mm}\texttt{COMPUTE increase = f1\_cd - f1\_sd} \\
\hspace*{12mm}\texttt{IF Transition is "Cross-Domain$\rightarrow$Multi-Domain"} \\
\hspace*{16mm}\texttt{COMPUTE increase = f1\_md - f1\_cd} \\
\hspace*{12mm}\texttt{IF increase > 0} \\
\hspace*{16mm}\texttt{COMPUTE increment positive\_count by 1} \\
\texttt{RETURN positive\_count} \\[1mm]
\textit{$L_{plan}=13$, $I_{retr}=6$, $I_{calc}=3$, $C_{flow}=10$, $S_{tab}=3$, $S_{cell}=336$} \\
\bottomrule
\end{tabular}
\caption{\textbf{Complete SciTaRC instances across the complexity range (1 of 2).} Each block shows the question, gold answer, expert-authored structured reasoning plan, and the plan-derived complexity metrics defined in Section~\ref{sec:metrics}. Instances are ordered by plan horizon ($L_{plan}$).}
\label{tab:appendix_examples}
\end{table*}

\begin{table*}[p]
\centering
\small
\renewcommand{\arraystretch}{1.15}
\begin{tabular}{@{}p{0.96\textwidth}@{}}
\toprule
\textbf{Q:} On which test sets are bigger models (either model size or longer token windows) always at least as good as smaller models of the same kind? \quad \textbf{A:} HellaSwag, PiQA \\[1mm]
\texttt{SELECT all test sets} \\
\texttt{SELECT all model families} \\
\texttt{SELECT all token-window groups} \\
\texttt{LOOP for each test set} \\
\hspace*{4mm}\texttt{COMPUTE violation\_exists = false} \\
\hspace*{4mm}\texttt{LOOP for each model family} \\
\hspace*{8mm}\texttt{SELECT models in this family ordered from smallest to largest} \\
\hspace*{8mm}\texttt{LOOP for each adjacent model pair} \\
\hspace*{12mm}\texttt{IF larger\_model\_score < smaller\_model\_score on this test set} \\
\hspace*{16mm}\texttt{COMPUTE violation\_exists = true} \\
\hspace*{4mm}\texttt{LOOP for each token-window group} \\
\hspace*{8mm}\texttt{SELECT models in this group ordered from shortest to longest} \\
\hspace*{8mm}\texttt{LOOP for each adjacent model pair} \\
\hspace*{12mm}\texttt{IF longer\_window\_score < shorter\_window\_score on this test set} \\
\hspace*{16mm}\texttt{COMPUTE violation\_exists = true} \\
\hspace*{4mm}\texttt{IF violation\_exists = false} \\
\hspace*{8mm}\texttt{SELECT this test set as valid} \\
\texttt{RETURN valid test sets} \\[1mm]
\textit{$L_{plan}=18$, $I_{retr}=9$, $I_{calc}=3$, $C_{flow}=12$, $S_{tab}=1$, $S_{cell}=165$ \quad (highest $C_{flow}$ in SciTaRC)} \\
\midrule
\textbf{Q:} Which language is the hardest for the Qwen2-Audio model in terms of LLM-based accuracy? For which models is a different language the hardest? \quad \textbf{A:} Korean; different for GPT-4o-mini-Audio (Cantonese) and GPT-4o-mini (Russian) \\[1mm]
\texttt{SELECT Qwen2-Audio, all languages} \\
\texttt{LOOP for each language} \\
\hspace*{4mm}\texttt{SELECT LLM accuracy in all settings} \\
\hspace*{4mm}\texttt{COMPUTE average1} \\
\texttt{COMPUTE argmin1} \\[1mm]
\texttt{SELECT all models} \\
\texttt{LOOP for each model} \\
\hspace*{4mm}\texttt{SELECT all languages} \\
\hspace*{4mm}\texttt{LOOP for each language} \\
\hspace*{8mm}\texttt{SELECT LLM accuracy in all settings} \\
\hspace*{8mm}\texttt{COMPUTE average2} \\
\hspace*{4mm}\texttt{COMPUTE argmin2} \\[1mm]
\hspace*{4mm}\texttt{IF argmin2 is not the same as argmin1} \\
\hspace*{8mm}\texttt{COMPUTE add this model to list} \\[1mm]
\texttt{RETURN argmin1, list} \\[1mm]
\textit{$L_{plan}=15$, $I_{retr}=6$, $I_{calc}=5$, $C_{flow}=6$, $S_{tab}=1$, $S_{cell}=280$ \quad (the task in Figure~\ref{fig:example})} \\
\bottomrule
\end{tabular}
\caption{\textbf{Complete SciTaRC instances across the complexity range (2 of 2).} Continued from Table~\ref{tab:appendix_examples}.}
\label{tab:appendix_examples_cont}
\end{table*}

\section{Inference Prompts}
\label{sec:inference_prompts}

In the following templates, \textbf{[Table Data]} represents the serialized text of the relevant tables, and \textbf{[Question]} represents the target query.

\subsection{Direct Inference (No Plan)}
These prompts establish the baseline performance for generalist (CoT) and specialist (PoT) models.

\begin{table*}[h]
\centering
\small
\renewcommand{\arraystretch}{1.2}
\begin{tabular}{p{0.95\textwidth}}
\toprule
\textbf{Standard Chain-of-Thought (CoT)} \\
\midrule

You are a helpful science assistant who answers questions about information in tables.

\vspace{0.5em}
Here is the relevant tabular data: \\
\textbf{[Table Data]}

\vspace{0.5em}
You may think through the question step by step. Your final response should be "Answer:" followed by the answer.

\vspace{0.5em}
Question: \textbf{[Question]} \\
\midrule
\textbf{Standard Program-of-Thought (PoT)} \\
\midrule

Write Python code to analyze the table and answer the question.

\vspace{0.5em}
Table: \\
\textbf{[Table Data]}

\vspace{0.5em}
Question: \textbf{[Question]}

\vspace{0.5em}
Write Python code that prints the final answer. \\
\bottomrule
\end{tabular}
\caption{Baseline prompts for Language and Code execution without explicit planning.}
\label{tab:standard_prompts}
\end{table*}

\subsection{Plan Generation}
This prompt is used in the \textbf{Autonomous Planning} setting to generate the structured reasoning path.

\begin{table*}[h]
\centering
\small
\renewcommand{\arraystretch}{1.1}
\begin{tabular}{p{0.95\textwidth}}
\toprule
\textbf{Plan Generation Prompt} \\
\midrule

You are a helpful assistant that creates step-by-step plans for answering questions about tables.

\vspace{0.5em}
\textbf{Operations:} \\

- SELECT: Extract specific values or subsets from the table \\
- LOOP: Iterate over items (e.g., LOOP for each model) \\
- COMPUTE: Numerical operations (average, argmax, difference, etc.) \\
- IF: Conditional logic \\
- RETURN: State what to return for the final answer

\vspace{0.5em}
\textbf{Rules:} \\

- Use exact syntax: "LOOP for each", "SELECT", "COMPUTE", "IF", "RETURN" \\
- Indent after LOOP and IF \\
- Use actual entity names, not generic "row" or "column"

\vspace{0.5em}
\textbf{Example Plan:} \\

LOOP for each model \\
\hspace{4mm} SELECT accuracy on Dataset A \\
\hspace{4mm} SELECT accuracy on Dataset B \\
\hspace{4mm} COMPUTE average across datasets \\
COMPUTE best\_model = argmax(average) \\
RETURN model with highest average accuracy

\vspace{0.5em}
Table: \\
\textbf{[Table Data]}

\vspace{0.5em}
Question: \textbf{[Question]}

\vspace{0.5em}
Write a concise step-by-step plan using the operations above. Do not solve it, just write the plan: \\
\bottomrule
\end{tabular}
\caption{The prompt used to elicit structured plans.}
\label{tab:planning_prompt}
\end{table*}

\subsection{Plan-Guided Inference}
These prompts inject the generated (Autonomous) or ground-truth (Oracle) plan into the context to guide the model's execution.

\begin{table*}[h]
\centering
\small
\renewcommand{\arraystretch}{1.2}
\begin{tabular}{p{0.95\textwidth}}
\toprule
\textbf{Plan-Guided Chain-of-Thought} \\
\midrule

You are a helpful science assistant who answers questions about information in tables.

\vspace{0.5em}
Here is the relevant tabular data: \\
\textbf{[Table Data]}

\vspace{0.5em}
Here is a step-by-step plan to follow: \\
\textbf{[Plan]}

\vspace{0.5em}
\textbf{How to read the plan:} \\

- SELECT: Look up and extract the specified values from the table \\
- LOOP for each X: Repeat the indented steps for every X \\
- COMPUTE: Perform the calculation described \\
- IF: Only do the indented steps when the condition is true \\
- RETURN: This is the final answer to provide

\vspace{0.5em}
Follow this plan carefully step by step to answer the question. Your final response should be "Answer:" followed by the answer.

\vspace{0.5em}
Question: \textbf{[Question]} \\
\midrule
\textbf{Plan-Guided Program-of-Thought} \\
\midrule

Write Python code to analyze the table and answer the question.

\vspace{0.5em}
Table: \\
\textbf{[Table Data]}

\vspace{0.5em}
Here is a step-by-step plan to implement: \\
\textbf{[Plan]}

\vspace{0.5em}
\textbf{How to read the plan:} \\

[...Same reading guide as above...]

\vspace{0.5em}
Question: \textbf{[Question]}

\vspace{0.5em}
Write Python code that implements this plan and prints the final answer. \\
\bottomrule
\end{tabular}
\caption{Prompts for executing a provided reasoning plan (Self or Gold).}
\label{tab:plan_guided_prompts}
\end{table*}

\section{Extended Complexity Analysis}
\label{sec:appendix_scale}

\begin{figure*}[t]
    \centering
    \includegraphics[width=\linewidth]{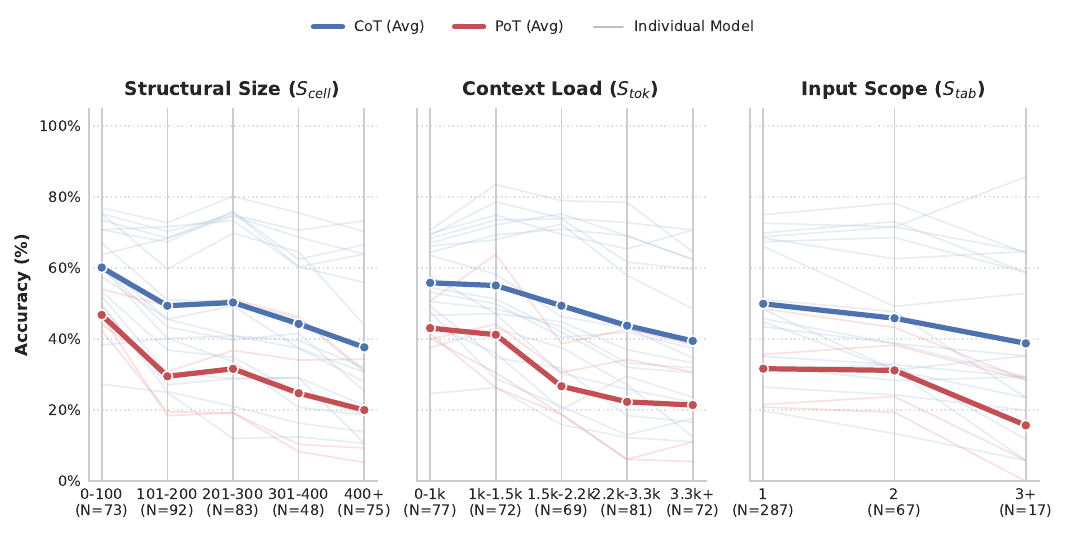}
    \caption{\textbf{Extended Complexity Analysis.} Performance breakdown across three dimensions of input scale: (Left) Structural Size ($S_{cell}$), (Middle) Context Load ($S_{tok}$), and (Right) Input Scope ($S_{tab}$). While increasing structural size and context load generally degrades performance for weaker models, multi-table reasoning shows divergent behaviors: top models often improve on tasks requiring cross-table linking. \textit{Sample sizes ($N$) for each bin are denoted on the x-axis.}}
    \label{fig:scale_appendix}
\end{figure*}

Figure~\ref{fig:scale_appendix} provides a comprehensive view of model performance against three dimensions of input scale: \textbf{Structural Size} ($S_{cell}$), \textbf{Context Load} ($S_{tok}$), and \textbf{Input Scope} ($S_{tab}$).

\textbf{Context Load ($S_{tok}$).}
Consistent with cell count, increasing the token length of the input exerts significant pressure on weaker models. Qwen3-8B drops by 30.0\% and Qwen3-Code by 36.0\% when moving from short ($<1k$) to long ($>3.3k$) contexts. However, reasoning-optimized frontier models remain immune to this noise, with DeepSeek-V3.2 (+1.5\%) and Grok-4.1-R (+1.1\%) showing slight performance gains in longer contexts.

\textbf{Input Scope ($S_{tab}$).}
Increasing the number of distinct tables (from 1 to 3+) reveals a divergence in capabilities. While typical models degrade, top reasoners like DeepSeek-V3.2 (+13.0\%) perform better on multi-table questions. This suggests that strong models possess the reasoning depth to handle the structured comparisons in these questions, successfully linking information across disjoint tables where weaker models lose context.

\section{Extended Planning Analysis}
\label{sec:appendix_planning}

\begin{figure*}[t]
    \centering
    \includegraphics[width=\linewidth]{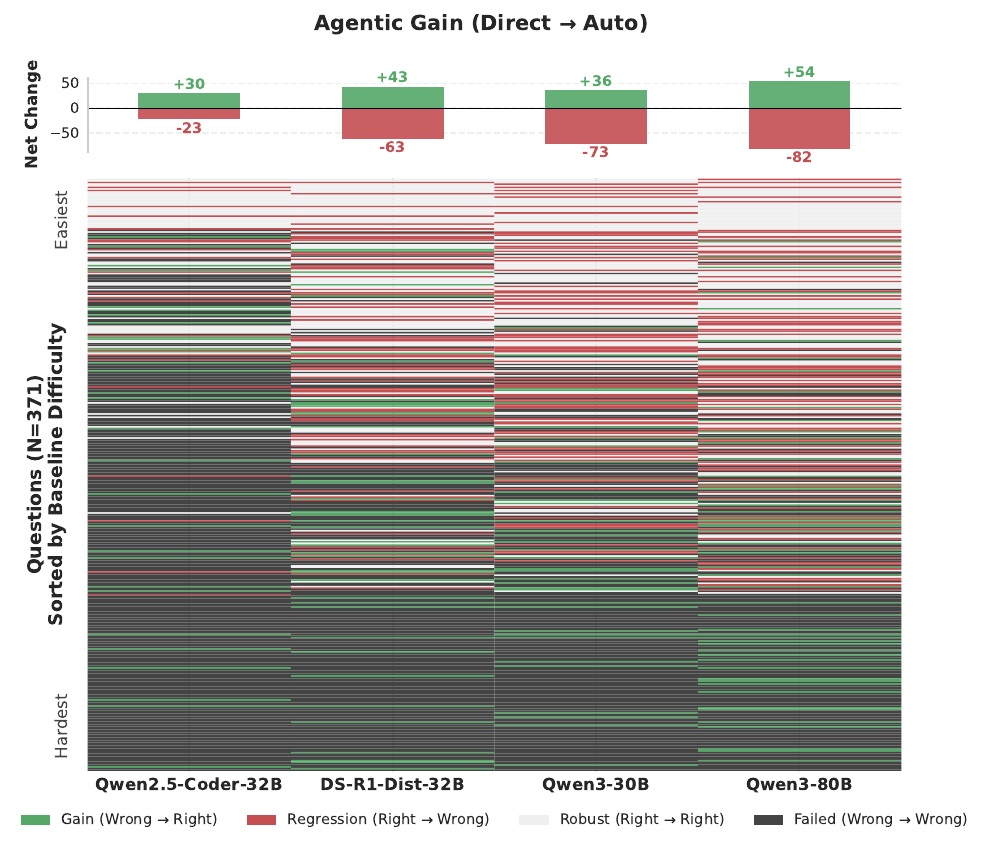}
    \caption{\textbf{Autonomous Gain (Direct $\to$ Autonomous).} Consistent with the Gain Curve, planning exhibits a dual effect. For the Code model (left), we observe net positive gains ($+7$) concentrated on harder questions. For Language models (right), we observe net regression (e.g., $-37$ for Qwen3-30B), primarily driven by the "Cost of Compliance" on easier questions where the overhead of planning disrupts simple retrieval tasks.}
    \label{fig:diff_no_auto}
\end{figure*}

\begin{figure*}[t]
    \centering
    \includegraphics[width=\linewidth]{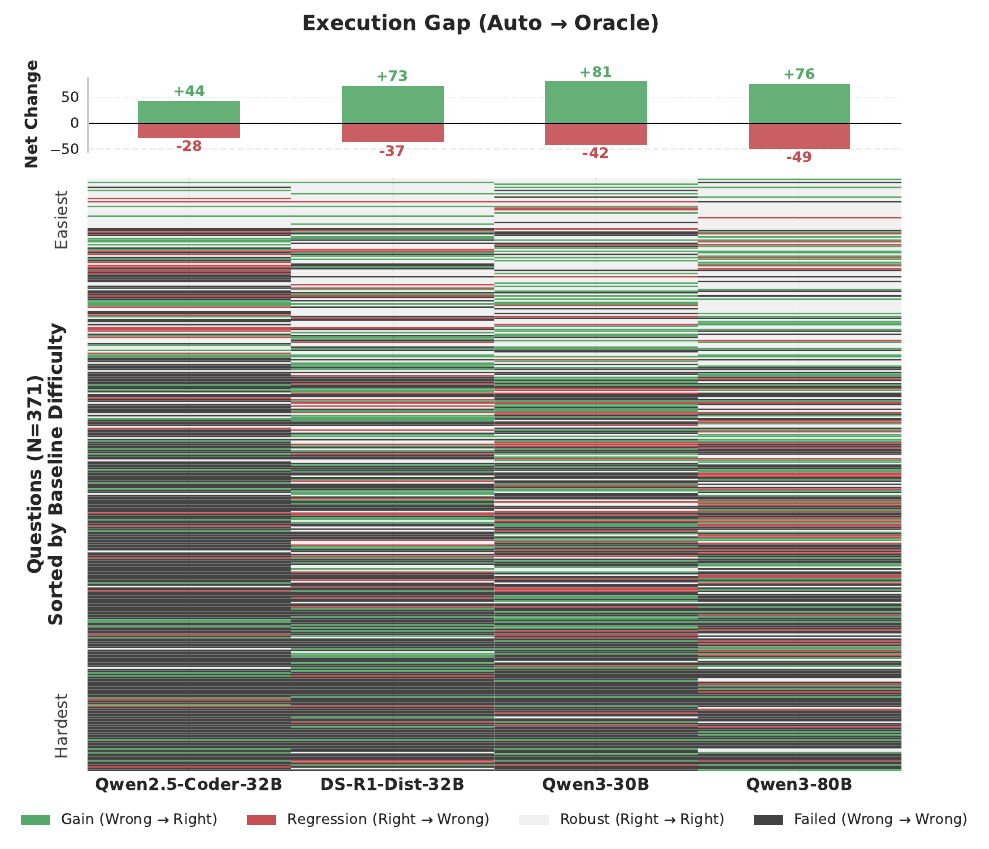}
    \caption{\textbf{Plan Correction (Autonomous $\to$ Oracle).} This plot isolates the quality of the generated plan. 
    While we observe net positive gains ($+16$ to $+39$) when moving to oracle plans, the non-trivial volume of \textbf{Regression (Red)} cases indicates that external strategies can sometimes disrupt the model's intrinsic reasoning flow. The dominance of shared failures (gray pixels) between autonomous and oracle planning confirms that the primary bottleneck is execution fidelity, as even perfect strategies often fail to overcome implementation errors.}
    \label{fig:diff_auto_oracle}
\end{figure*}

\begin{figure*}[t]
    \centering
    \includegraphics[width=\linewidth]{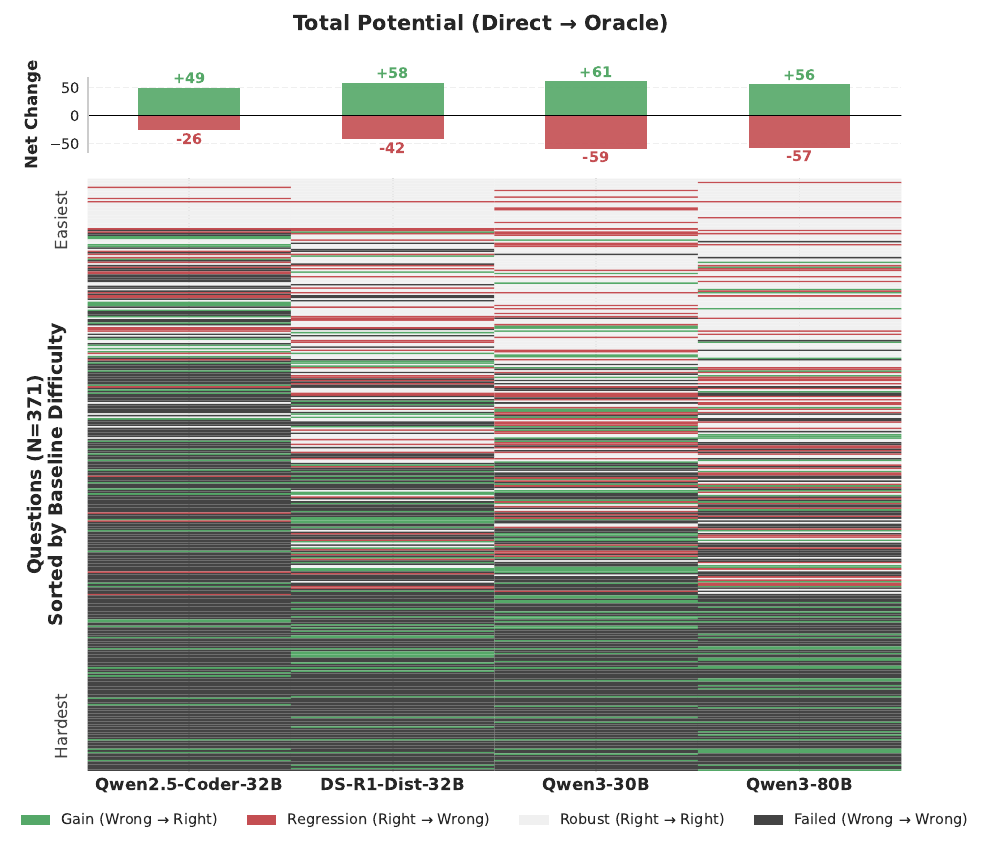}
    \caption{\textbf{Total Potential (Direct $\to$ Oracle).} This compares the baseline directly against the upper bound. Even with perfect Oracle guidance, the net gains are moderate ($-1$ to $+23$). The vast majority of difficult questions remain \textbf{Failed (Gray)}, demonstrating that a significant portion of the dataset hits a hard capability ceiling that cannot be solved by planning assistance alone.}
    \label{fig:diff_no_oracle}
\end{figure*}

To complement the probabilistic Gain Curves, we present differential heatmaps ($N=371$) in Figures~\ref{fig:diff_no_auto}--\ref{fig:diff_no_oracle}. These plots visualize the exact state changes for individual questions, distinguishing between "Gain" (Wrong $\to$ Right) and "Regression" (Right $\to$ Wrong).

\section{Few-Shot Ablation}
\label{sec:appendix_fewshot}

The main experiments evaluate all models zero-shot. To test whether in-context demonstrations can teach models the structured plan syntax and thereby relieve the execution bottleneck, we ran an additional 2-shot ablation on four open-weight models. We prepended two demonstrations to the prompt, each consisting of a table, a question, a step-by-step reasoning trace (CoT) or execution trace (PoT), the final answer, and, in the \textbf{FS-Oracle} condition, the corresponding structured plan. The two demonstration questions are held out, so all few-shot numbers are computed over the remaining 369 questions; this smaller denominator accounts for the minor differences between the ZS-Direct column below and Table~\ref{tab:main_results}. Proprietary models are omitted because of API constraints on long prompts.

\begin{table}[h]
\centering
\small
\renewcommand{\arraystretch}{1.1}
\begin{tabular*}{\linewidth}{@{\extracolsep{\fill}} l ccc | cc @{}}
\toprule
\multirow{2}{*}{\textbf{Model}} & \multicolumn{3}{c}{\textbf{Accuracy (\%)}} & \multicolumn{2}{c}{\textbf{FS-Oracle Errors}} \\
\cmidrule(lr){2-4} \cmidrule(lr){5-6}
& \textbf{ZS-Direct} & \textbf{FS-Direct} & \textbf{FS-Oracle} & \textbf{Total} & \textbf{Regressions} \\
\midrule
Qwen3-30B              & 44.4 & 47.1 & \textbf{47.7} & 193 & 52 \\
Gemma-3-27B            & 30.1 & 28.7 & \textbf{32.5} & 249 & 57 \\
Llama-3.3-70B          & 34.4 & 29.0 & \textbf{29.3} & 261 & 71 \\
Qwen3-Coder-30B (PoT)  & 19.5 & 25.7 & \textbf{35.2} & 239 & 18 \\
\bottomrule
\end{tabular*}
\caption{\textbf{Two-shot ablation ($N=369$).} \textbf{ZS-Direct}: zero-shot, no guidance. \textbf{FS-Direct}: two demonstrations. \textbf{FS-Oracle}: two demonstrations plus the ground-truth structured plan. \textbf{Regressions} counts questions answered correctly under ZS-Direct but failed under FS-Oracle.}
\label{tab:fewshot}
\end{table}

Two observations carry over from the zero-shot setting. First, the ceiling remains low. Even with demonstrations \textit{and} a correct plan, the strongest model reaches 47.7\%, failing more than half the benchmark. Teaching the plan syntax explicitly does not convert correct strategies into correct answers, which is consistent with execution rather than comprehension being the limiting factor.

Second, the compliance cost persists and remains architecture-dependent. The code model gains the most from oracle plans (+15.7 points over ZS-Direct) and regresses on only 18 questions. The generalist models gain little or lose ground, and 52--71 of their previously correct answers turn wrong when they are required to follow an external plan. Demonstrations do not eliminate this cost; the tension between a rigid algorithmic path and a generalist model's intrinsic reasoning survives explicit instruction in the format.

\end{document}